# KNN Classification With One-Step Computation

Shichao Zhang, *Senior Member, IEEE* and Jiaye Li

**Abstract**—KNN classification is an improvisational learning mode, in which they are carried out only when a test data is predicted that set a suitable K value and search the K nearest neighbors from the whole training sample space, referred them to the lazy part of KNN classification. This lazy part has been the bottleneck problem of applying KNN classification due to the complete search of K nearest neighbors. In this paper, a one-step computation is proposed to replace the lazy part of KNN classification. The one-step computation actually transforms the lazy part to a matrix computation as follows. Given a test data, training samples are first applied to fit the test data with the least squares loss function. And then, a relationship matrix is generated by weighting all training samples according to their influence on the test data. Finally, a group lasso is employed to perform sparse learning of the relationship matrix. In this way, setting K value and searching K nearest neighbors are both integrated to a unified computation. In addition, a new classification rule is proposed for improving the performance of one-step KNN classification. The proposed approach is experimentally evaluated, and demonstrated that the one-step KNN classification is efficient and promising.

## 1 Introduction

KNN classification is one of top 10 algorithms in data mining [1]. It has widely and successfully been applied to many areas, such as text classification [2], pattern recognition [3] and image processing [4]. Standard KNN classification predicts a test data with only its K nearest neighbors in the training dataset. One need only select/construct a distance function and a classification rule when designing his/her KNN classification algorithm [5]. A KNN classification algorithm is a query triggered yet improvisational learning procedure, in which they are carried out only when a test data is predicted that set a suitable K value and search the K nearest neighbors from the whole training sample space. This means KNN classification includes a lazy part, setting K value and searching K nearest neighbors. From the learning procedure, designing a KNN classification algorithm is almost independent of the training data, i.e., it does not train anything with the training dataset before a query is submitted. Therefore, KNN classification is attributed to a non-model, or nonparametric, or lazy classification method [6].

To deal with the lazy part of KNN classification, there are recently many efforts to setting K value or searching K nearest neighbors. For example, Gora and Wojna presented an algorithm of learning the optimal-K-values for a test data [7]. Guo *et al.* advocated to automatically seek the optimal-K-value for a test data [8]. Wang *et al.* argued to locally adjust the number of nearest neighbors [9]. Manocha *et al.* designed a probabilistic method for computing the optimal-K-values of nearest neighbors [10]. Li *et al.* spelled out why different categories need different numbers of nearest neighbors [11]. Sun *et al.* designed an adaptive algorithm of finding the optimal-K-values for a test data [12]. Zhang *et al.* advocated to reconstruct the relation matrix and built a optimal-K-value computation [5]. Gou *et al.* proposed a K-nearest neighbor algorithm by local mean vector representation [13]. Pan *et al.* proposed an adaptive KNN algorithm, which not only considers the majority class, but also considers the second majority class in the K nearest neighbors of a test data [14].

However, the above KNN classification algorithms are still lazy. Recently, a non-lazy KNN classification has been developed in [15], denoted as Ktree and K*tree classification algorithms. Ktree is designed for fast finding the K value of a test data. It adds a training stage to traditional KNN method and thus outputting a training model, i.e., building a decision tree, Ktree, to predict the optimal-K-values for each test data. K*tree is proposed for both setting K value or searching K nearest neighbors. It is an extension of Ktree by appending sets of nearest neighbors of training samples in each leaf node. In this way, the K*tree KNN classification is not lazy.

From these reports, setting K value and searching K nearest neighbors are carried out in two phases. In general, searching K nearest neighbors is a complete training sample space search, i.e., it is an NP problem [16]. Although the K*tree KNN classification is not lazy, the lazy part of KNN classification is also accomplished in two steps as follows. For a test data, the first step searches its nearest neighbor X and the K value. The second step identifies the K-1 nearest neighbors from the set of nearest neighbors of X.

To attack the above issues in KNN classification, a one-step KNN classification algorithm is designed in this article that reduces the complexity of searching K nearest neighbors to $O(N^2)$. The one-step KNN classification is a one-step computation that integrates both setting K value and searching K nearest neighbors to a unified computation. Specifically, given a test data, training samples are first applied to fit the test data with the least squares loss function. And then, a relationship matrix is generated by

• *The authors are with the School of Computer Science and Engineering, Central South University, Changsha 410083, China. E-mail: {zhangsc, lijiaye}@csu.edu.cn.*

*Manuscript received 21 Dec. 2020; revised 23 Aug. 2021; accepted 6 Oct. 2021. Date of publication 0 . 0000; date of current version 0 . 0000.*
*This work was supported in part by the Natural Science Foundation of China under Grant 61836016.*
*(Corresponding author: Jiaye Li.)*

weighting all training samples according to their influence on the test data. Finally, a group lasso is employed to perform sparse learning of the relationship matrix. In this way, setting K value and searching K nearest neighbors are both integrated to a unified computation. In addition, a new classification rule is proposed for improving the performance of one-step KNN classification. The proposed approach is experimentally evaluated, and demonstrated that the one-step KNN classification is efficient and promising.

The key work in this paper is summarized as follows.

- A new non-lazy KNN algorithm is proposed, i.e., it can learn the optimal K value and K neighbors of the test data in the training process.
- The lazy part of KNN is calculated in one step by matrix calculation, i.e., setting K value and searching K nearest neighbors are both done at the same step, which is different from K*tree (K*tree requires two steps) in [15].
- We conducted a series of experiments to show that the proposed algorithm exceeds the state-of-the-art methods in terms of classification performance and running cost.

The rest of this article is organized as follows. Section 2 briefly reviews previous related work for KNN. Section 3 describes in detail our proposed method, the optimization process and the proof of convergence. Section 4 shows the results of all algorithms on simulated datasets and real datasets. Section 5 summarizes the full paper.

## 2 Preliminary

This section briefly recalls some main reports on setting K value and searching K nearest neighbors. In addition, we also briefly introduce the application of KNN in practice.

### 2.1 K Value Calculation

The KNN classification algorithm was first proposed by Cover and Hart in 1967 [17]. After a long period of research, researchers have produced some improved KNN algorithms [18], [19]. As we all know, in KNN, K value calculation and nearest neighbor query are two very critical issues. Among them, the nearest neighbor query can be solved by different distance measurement functions, such as (euclidean distance, Mahalanobis distance, Manhattan distance and angle cosine distance, *etc.* ). For K value calculation, the current mainstream method is through expert settings or cross-validation methods. They do not fundamentally solve the problem of K value calculation because all test data have the same K value. If the K value is selected too small, it will easily cause over-fitting. Choosing a too large K value will increase the approximate error of learning [20]. Therefore, Zhang *et al.* advocated to reconstruct the relation matrix and built a optimal-K-value computation [5]. Pan *et al.* proposed an adaptive KNN algorithm [14], which not only considers the majority class, but also considers the second majority class in the K nearest neighbors of a test data . And it also performs automatic K value selection by sorting the distance between the test data and the discriminant center. Sun and Huang also proposed an adaptive K-nearest neighbor algorithm [12]. It first calculates the optimal K value of each training data according to the euclidean distance. And then, it finds the nearest neighbor of the test data, and uses the optimal K value of the nearest neighbor as the optimal K value of the test data. Finally, the KNN classification is carried out according to the majority rule. Li *et al.* set different neighbors for different classes [21]. The large class is set a larger K value, and the small class is set a smaller K value. It makes KNN more suitable for text classification. Liu *et al.* proposed a weighted KNN algorithm, which obtains the local K value of the test data according to the training data, and then calculates the weight of each neighbor according to the distance. Finally, it gets the class label of the test data according to majority rule [22]. Mahin *et al.* set the K value through Geometric Mean (G-mean), and applied it to the unbalanced datasets [23]. Assegie selected the optimal K value based on the misclassification rate, and applied the proposed KNN algorithm to breast cancer detection [24].

Zhang *et al.* proposed an effective KNN algorithm, which uses training data to reconstruct training data to obtain the optimal K value of all training samples [15]. Then calculate the optimal K value of the test data by constructing Ktree and K*tree. The difference is that K*tree effectively reduces the calculation amount of the algorithm, which greatly reduces the running time of the algorithm test. In the Ktree, the leaf nodes of Ktree store the optimal K value of all training data. Given a test data, we determine its optimal K value based on Ktree discrimination. After obtaining the optimal K value, we determine the class label of the test data from all the training data according to the majority rule. It can be seen that the Ktree algorithm needs to find K neighbors from all the training data, and its search range is the entire training set, which takes a lot of time. Therefore, K*tree was proposed, and the difference between it and ktree lies in the leaf nodes. The leaf nodes of K*tree not only store the optimal K value, but also store the training samples with the same optimal K value, the nearest neighbor of the training sample with the same optimal K value, and the K nearest neighbor of nearest neighbor. In this way, when looking for the K-nearest neighbors of the test data, we only need to look for the corresponding leaf nodes. This means the search range of K*tree is leaf nodes (including training data with the same K value, their nearest neighbors, and K nearest neighbors of their nearest neighbors), which is a relatively small subset compared to the entire training set. Therefore, the running time of K*tree is greatly reduced compared to Ktree.

### 2.2 Neighbor Search

In the nearest neighbor search, researchers have done a lot of work in this area, and they improve KNN by proposing new distance measurement functions.

Abu-Aisheh proposed a multi-graph edit distance for KNN, the algorithm achieved faster time especially when the number of graphs is large [25]. Gou *et al.* proposed a KNN algorithm by local mean representation [13]. In the algorithm, a new distance function is constructed with the linear representation between test data and training data. This function is based on the relationship between training data and test data. Jiao *et al.* proposed a paired distance metric to improve the KNN algorithm, which is different from other methods to learn a global distance metric [26]. Xie *et al.* used spearman distance to improve the KNN algorithm and apply it to indoor positioning [27]. Manal

Marzouq proposed a new distance-weighted KNN based on the kernel method, and applied it to global solar radiation estimation [28]. Wang *et al.* proposed a locally adaptive distance to solve the problem that different types of patterns in KNN overlap in certain areas of the feature space [29]. Yu *et al.* defined an iDistance, which enables KNN to find a suitable K-nearest neighbor in a high-dimensional space. In addition, it still uses the $B_{+-}$ tree structure to index the data [30]. Karthick Ganesan and Harikumar Rajaguru studied the impact of various distance functions on KNN classification and applied it to MRI Images [31]. Tehrany *et al.* improved KNN based on manhattan distance metric. It can be applied to the android system to detect malicious installers [32]. Candra Dewi and Yoke Kusuma Arbawa studied the influence of the distance function in KNN on the classification of soil organic matter [33]. Samiee *et al.* proposed a low-rank local distance function to improve the KNN algorithm, which solves the limitation of using the same distance metric for global data. In addition, it can take into account non-linear relationships in the data [34].

In addition to the improvement of the distance function to perform the nearest neighbor query, some works have also improved the distance function to enable KNN to be applied to different cases. Santos *et al.* studied the influence of different distance functions on KNN in imputation of missing data [35]. Tharmakulasingam *et al.* used non-euclidean distance and KNN to improve pathogen identification [36]. Wang *et al.* proposed a new KNN algorithm through similarity and location distance [37]. Specifically, it first proposes a new weighted euclidean distance metric according to the attenuation law of spatial signals. Then it chooses an appropriate distance between the location distance and the weighted euclidean distance. Finally, the user's location is estimated by the proposed weighted KNN algorithm. Geler *et al.* proposed a constrained elastic distance for KNN and applied the KNN algorithm to time series classification [38].

Zhang *et al.* proposed an ML-KNN algorithm for multi-label learning [39]. It first finds K neighbors in the training set. Then the statistical information of the label in these neighbors is obtained. Finally, the maximum a posteriori principle is used to determine the label set of the test data. This method is still a lazy learning method. Zhang *et al.* proposed a KNN method and applied it to unbalanced data distribution [40]. Sitawarin *et al.* verified the robustness of deep K-nearest neighbors. It combines deep learning and KNN to improve the robustness of adversarial examples [41]. DJENOURI *et al.* proposed an adaptive K nearest neighbor algorithm [42]. Specifically, it first considers space and time information. Then it builds a database from the traffic flow data, and applies anomaly detection mechanism to the flow distribution probability. Finally, it uses the distance-based KNN algorithm to detect the true distribution of outliers. Salvador-Meneses *et al.* proposed a compressed KNN algorithm, which solves the problem that KNN is not suitable for big data [43]. Specifically, it first compresses the data into different groups, and then stores a certain amount of features in each group to avoid the problem of memory overflow. Finally, in the experiment, it reduces the running time of the algorithm. Shi *et al.* proposed an enhanced K-nearest neighbor algorithm [44]. This method embeds the dimensionality reduction algorithm into the extraction of sensitive information, and applies the enhanced KNN algorithm to tremor recognition under different conditions. In addition, the effectiveness of this method is verified by experiments. Rachdi *et al.* proposed a Knn local linear regression algorithm [45]. It combines KNN and local linear estimation and inherits many advantages of these two methods. Finally, its effectiveness was verified on simulated data and real datasets. Bian *et al.* proposed an adaptive fuzzy KNN algorithm [46]. Specifically, in the training process, it learns the optimal K value of each test data after reconstruction through sparse learning. In addition, it also builds a decision tree to improve the running speed of the proposed KNN algorithm. The research of this method can be regarded as based on the research of Zhang *et al.* [15]. Giri *et al.* applied the Knn algorithm to a position prediction system [47]. Dian *et al.* proposed a progressive top-KNN search algorithm [48]. This method optimizes the index structure, thereby improving the efficiency of index maintenance in large networks. Yan *et al.* proposed a KNN search algorithm through random projection forest [49]. It finds neighbors by combining multiple KNN sensitive trees. Each tree is constructed by random projection. The experiment also shows the superiority of the algorithm.

Le *et al.* proposed a K-nearest neighbor algorithm with deep similarity [50]. It applies a new similarity function to map data to high-dimensional space, which makes the accuracy of KNN in high-dimensional space higher. It can dig into the nonlinear relationship in the data. Gottlieb proposed a sample compression algorithm based on nearest neighbors [51]. Zhu *et al.* proposed an adaptive Knn algorithm, which mainly deals with the non-linear relationship in the data, insufficient training data and time changes [52]. It searches for the target prototype through iterative update distance metrics, and finally applies to process monitoring. M. Ahmed *et al.* proposed an optimized KNN algorithm, which uses an adaptive function to select the best location of neighbor sensors to improve the life of the wireless sensor network [53].

Compared with setting the K value, the researcher does more work on searching K nearest neighbors. But work that combines the two is rarely seen. Setting the K value and searching K nearest neighbors through one-step calculation is a work that has not been done before.

## 3 Our Approach

### 3.1 Notation

In this article, we use uppercase bold letters to represent matrices and lowercase bold letters to represent vectors. Training data matrix $\mathbf{X} \in \mathbb{R}^{n\times d}$. Test data matrix $\mathbf{Y} \in \mathbb{R}^{m\times d}$. $\mathbf{X}^i$ and $\mathbf{X}_j$ represent the $i$th row and $j$th column of $\mathbf{X}$, respectively. $X_{i,j}$ represents the $j$th element of the $i$th row in the matrix $\mathbf{X}$. The $l_f$-norm and $l_1$-norm of $\mathbf{X}$ are denoted as $\|\mathbf{X}\|_F = \sqrt{\sum_i^n \|x^i\|_2^2} = \sqrt{\sum_j^d \|x_j\|_2^2}$ and $\|\mathbf{x}\|_1 = \sum_{i=1}^n |x_i|$ respectively. The transpose, inverse and trace of matrix $\mathbf{X}$ are denoted as $\mathbf{X}^T$, $\mathbf{X}^{-1}$, and $tr(\mathbf{X})$ respectively.

We summarize these notations used in our paper in Table 1.

### 3.2 Framework

In this section, we elaborate on our proposed one-step KNN algorithm. Specifically, we first introduce the proposed objective function to obtain the optimal K value of each test data, K nearest neighbors of each test data and the weights of the nearest neighbors, and then use a weighted classification rule to perform KNN classification. Fig. 1 shows the detailed process of our algorithm.

TABLE 1
The Detail of the Notations Used in This Paper

| | |
|---|---|
| $\mathbf{X}$ | training data matrix |
| $\mathbf{Y}$ | test data matrix |
| $\mathbf{X}^i$ | the $i$th row of $\mathbf{X}$ |
| $\mathbf{X}_j$ | the $j$th column of $\mathbf{X}$ |
| $X_{i,j}$ | the $j$th element of the $i$th row in the $\mathbf{X}$ |
| $\|\mathbf{X}\|_F$ | the frobenius norm of $\mathbf{X}$, i.e., $\|\mathbf{X}\|_F = \sqrt{\sum_{i,j} \mathbf{x}_{i,j}^2}$ |
| $\|\mathbf{x}\|_1$ | the $l_1$ -norm of $\mathbf{X}$, i.e., $\|\mathbf{x}\|_1 = \sum_{i=1}^{n} \|x_i\|$ |
| $\|\mathbf{x}\|_0$ | the $l_0$ -norm of $\mathbf{X}$, i.e., the number of non-zero elements in $\mathbf{x}$ |
| $\mathbf{X}^T$ | the transpose of $\mathbf{X}$ |
| $\mathbf{X}^{-1}$ | the inverse of $\mathbf{X}$ |
| $tr(\mathbf{X})$ | the trace of $\mathbf{X}$ |

### 3.2.1 Sparse Model Based on Linear Representation

In sparse learning, researchers usually build sparse models in the form of loss function + sparse regular term. It can learn the inherent sparse expression of the relationship between data. The traditional objective function is as follows:

$$\min_{\mathbf{W}} f(\mathbf{X}, \mathbf{W}) + \varphi(\mathbf{W}), \tag{1}$$

where $f(\mathbf{X}, \mathbf{W})$ is the loss function, which can be any loss function, such as least square loss, cross entropy loss, and hinge loss. $\varphi(\mathbf{W})$ is a sparse regular term. Common sparse regular terms include $l_1$-norm, $l_2$-norm and $l_{2,p}$-norm. By solving the above formula, the optimal solution of W can be finally obtained for feature selection or other learning tasks.

In this paper, we use the least squares loss function due to its high fit. As shown in the following formula:

$$\min_{\mathbf{W}} \|\mathbf{Y} - \mathbf{X}\mathbf{W}\|_F^2 + \varphi(\mathbf{W}). \tag{2}$$

In the above formula, $\mathbf{Y}$ represents the test sample and $\mathbf{X}$ represents the training sample. In practice, in order to avoid the invertible $\mathbf{W}$ matrix, and to achieve the effect of sparseness, people usually add a sparse regular term at the back. In this paper, we use the improved group lasso for sparse restriction, as shown in the following formula.

$$\varphi(\mathbf{W}) = \sum_{g \in G} \|\mathbf{W}_{Gg}\|_1^2, \tag{3}$$

where $G$ represents the set of all groups, and $g$ represents the $g$th group. In the group lasso of this article, we group all the training data, and then make the intra group sparse and inter group non sparse. This can effectively consider the similarity relationship within the data. So as to better find the optimal K value and corresponding neighbors of the test data. In addition, we also impose a weight on all training data, which can effectively measure the importance of each neighbor. We get the final objective function as follows.

$$\begin{aligned} &\min_{\mathbf{W}} \left\|\mathbf{Y}^T - \mathbf{X}^T \Theta \mathbf{W}\right\|_F^2 + \alpha\|\mathbf{W}\|_F^2 + \beta\|\mathbf{W}\|_1^2 \\ &st.\ \theta > 0, \mathbf{1}^T\theta = 1, \end{aligned} \tag{4}$$

where $\Theta \in \mathbb{R}^{n \times n}$ is a diagonal matrix, the diagonal elements are the corresponding elements in $\theta$, and the value of each element represents the weight of the training data. When $\alpha = 0$, Formula (4) can be regarded as a group lasso. $\|\mathbf{W}\|_1^2$ is the group lasso sparse regular term. $\mathbf{W} \in \mathbb{R}^{n \times m}$ is the relational matrix, which contains the optimal K value of each test data and the corresponding K nearest neighbors. For example, the final solution of $\mathbf{W}$ is as follows.

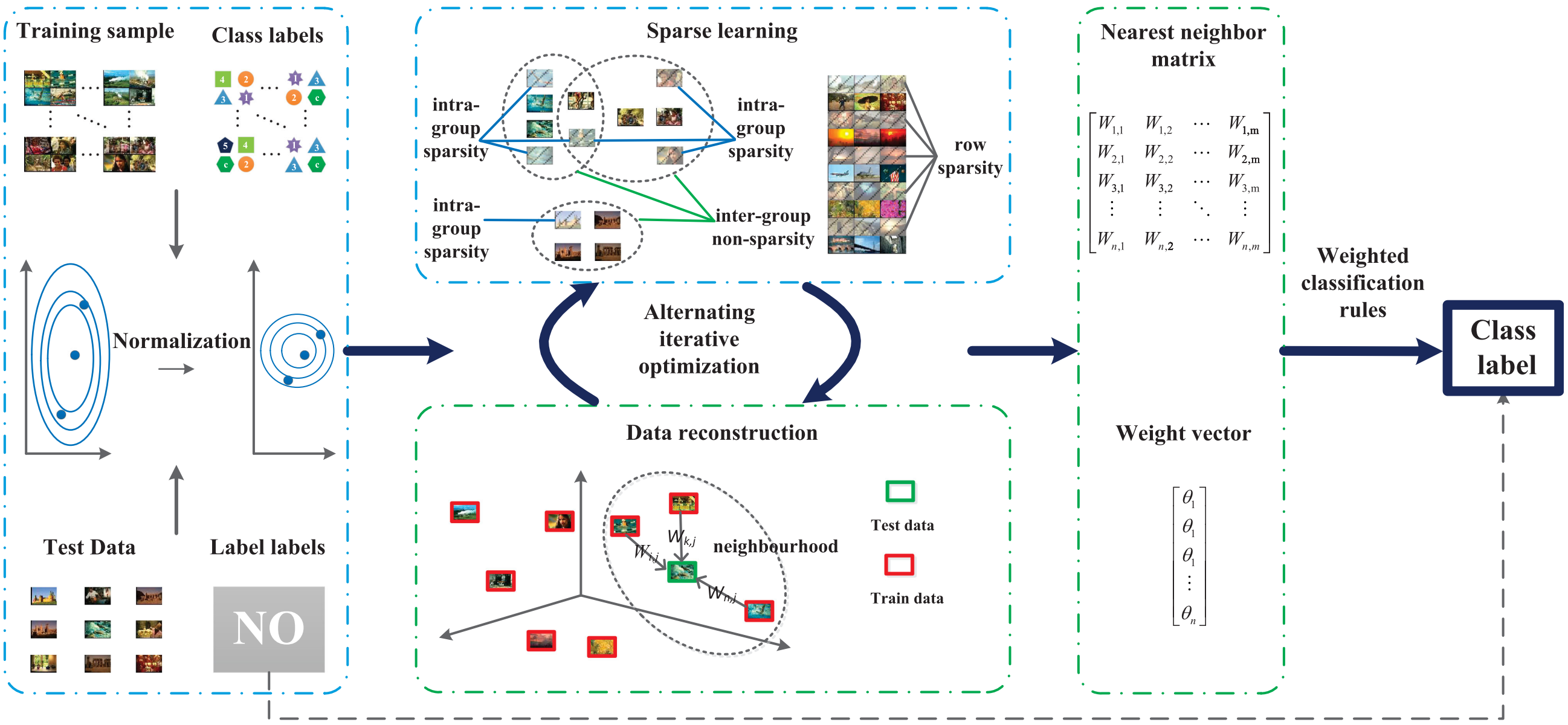

Fig. 1. Flow chart of the proposed method.

$$\mathbf{W} = \begin{bmatrix} 1.2 & 2.1 & 1.4 & 0 \\ 0.6 & 0 & 1.3 & 1.2 \\ 0 & 1.2 & 1.1 & 0 \\ 0 & 0 & 1.1 & 0 \\ 1.2 & 2.1 & 0 & 1.2 \end{bmatrix}. \tag{5}$$

In the above formula, $\mathbf{W} \in \mathbb{R}^{5\times4}$. It represents the relationship between 4 test data and 5 training data. The number of non-zero elements in each column of the matrix $\mathbf{W}$ is the optimal K value of the test data. For example, in the above $\mathbf{W}$, there are 3 non-zero elements in the first column, the optimal K value of the first test data is 3, and its neighbors are the 1st, 2rd and 5th training samples. Similarly, the optimal K values and corresponding nearest neighbors of the 2nd, 3rd and 4th test data are: 3 (1st, 3rd and 5th training data), 4 (1st, 2nd, 3rd and 4th training data), and 2 (2nd and 5th training data). In this way, we can get the optimal K value and K nearest neighbors for each test data.

Next, we proposed a new classification rule, as follows.

$$v(\mathbf{y}_i) = \arg\max_{c\in\Omega} \frac{1}{\|\mathbf{W}_j\|_0} \sum_{c\in\Omega} \Theta\mathbf{W}_j I(l = c). \tag{6}$$

Equation (6) can be regarded as a new classification rule. $\|\mathbf{W}_j\|_0$ represents the number of non-zero elements in the $j$th column (i.e., the optimal K value of the $j$th test data). And $\Theta$ is the weight of all training data. When it is multiplied by $\mathbf{W}_j$, it is equivalent to weighting only the neighbors, because the $\mathbf{W}$ element value corresponding to non-nearest neighbors is zero. Then, the predicted class label of test data $\mathbf{y}_i$ can be directly obtained by Formula (6). $I(\bullet)$ is the indicator function. It means: $I(\bullet) = 1$, if the content in parentheses is true; $I(\bullet) = 0$, otherwise.

### 3.3 Optimization

In this part, we optimize and solve the proposed objective function. We first transform Formula (4) through the following theory.

**Theorem 1.** *Formula (4) is equivalent to the following formula.*

$$\min_{\mathbf{W}} \|\mathbf{Y}^T - \mathbf{X}^T\tilde{\mathbf{W}}\|_F^2 + \alpha\|\tilde{\mathbf{W}}\|_{2,1}^2 + \beta\|\tilde{\mathbf{W}}\|_1^2, \tag{7}$$

*where $\theta$ can be obtained by the following formula:*

$$\theta_j = \frac{\left\|\tilde{\mathbf{W}}^j\right\|_2 + \left\|\tilde{\mathbf{W}}^j\right\|_1}{\sum_{j=1}^n \left\|\tilde{\mathbf{W}}^j\right\|_2 + \sum_{j=1}^n \left\|\tilde{\mathbf{W}}^j\right\|_2}. \tag{8}$$

**Proof.** Let $\tilde{\mathbf{W}} = \Theta\mathbf{W}$. Then $\mathbf{W} = \Theta^{-1}\tilde{\mathbf{W}}$. The Formula (4) can be written as follows:

$$\min_{\mathbf{W}} \|\mathbf{Y}^T - \mathbf{X}^T\tilde{\mathbf{W}}\|_F^2 + \alpha\sum_{j=1}^n \frac{\left\|\tilde{\mathbf{W}}^j\right\|_2^2}{\theta^j} + \beta\sum_{j=1}^n \frac{\left\|\tilde{\mathbf{W}}^j\right\|_1^2}{\theta^j} \tag{9}$$

$s.t., \theta > 0, \mathbf{1}^T\theta = 1.$

If $\tilde{\mathbf{W}}$ is fixed, we can get the optimal solution of $\theta$ by solving the following equation.

$$\min_{\theta>0, \mathbf{1}^T\theta=1} \alpha\sum_{j=1}^n \frac{\left\|\tilde{\mathbf{W}}^j\right\|_2^2}{\theta^j} + \beta\sum_{j=1}^n \frac{\left\|\tilde{\mathbf{W}}^j\right\|_1^2}{\theta^j}. \tag{10}$$

It can be verified that the optimal solution of $\theta$ is

$$\theta^j = \frac{\left\|\tilde{\mathbf{W}}^j\right\|_2 + \left\|\tilde{\mathbf{W}}^j\right\|_1}{\sum_{j'=1}^n \left\|\tilde{\mathbf{W}}^j\right\|_2 + \sum_{j'=1}^n \left\|\tilde{\mathbf{W}}^j\right\|_1}. \tag{11}$$

Through Formula (11), Formula (10) can be written as follows.

$$\min_{\theta>0, \mathbf{1}^T\theta=1} \alpha\|\tilde{\mathbf{W}}\|_{2,1}^2 + \beta\|\tilde{\mathbf{W}}\|_1^2. \tag{12}$$

Therefore, Formula (4) can be rewritten as

$$\min_{\tilde{\mathbf{W}}} \|\mathbf{Y}^T - \mathbf{X}^T\tilde{\mathbf{W}}\|_F^2 + \alpha\|\tilde{\mathbf{W}}\|_{2,1}^2 + \beta\|\tilde{\mathbf{W}}\|_1^2. \tag{13}$$

□

When $\alpha = 0$, Formula (13) can be regarded as a group lasso. In group lasso, all samples are grouped, and the samples in each group are very similar. The samples between groups are not similar. It uses $l_1$ -norm for sparseness within groups, and $l_2$ -norm for non-sparseness between groups. It can effectively choose the neighbor samples of the test data from all the training data, and get the optimal K value. Equation (13) can be equivalent to the following equation.

$$\min_{\tilde{\mathbf{W}}} \|\mathbf{Y}^T - \mathbf{X}^T\tilde{\mathbf{W}}\|_F^2 + \alpha tr(\tilde{\mathbf{W}}^T\mathbf{Q}\tilde{\mathbf{W}}) + \beta tr(\tilde{\mathbf{W}}^T\mathbf{F}\tilde{\mathbf{W}}), \tag{14}$$

where $\mathbf{Q}$ and $\mathbf{F}$ are diagonal matrices, and their diagonal elements are as follows.

$$q_{jj} = \frac{\sum_{v=1}^n \sqrt{\|\tilde{\mathbf{W}}^v\|_2^2}}{\sqrt{\left\|\tilde{\mathbf{W}}^j\right\|_2^2}} \tag{15}$$

$$f_{ii} = \sum_{g\in\Omega} \frac{(\mathbf{I}_{Gg})_i \|\tilde{\mathbf{W}}_{Gg}\|_1}{\left\|\tilde{\mathbf{W}}^i\right\|_1} (i = 1, \ldots, n), \tag{16}$$

where $\Omega$ represents all groups. $\mathbf{I}_{Gg} \in \mathbb{R}^{n\times1}$, The elements in $\mathbf{I}_{Gg}$ can only be 0 or 1. i.e., if there are $j$ samples in the group, the element value of the corresponding position is 1, and the element value of other positions is 0. For example, $G_1 = \{1, 2, 4, 6\}$, the corresponding $\mathbf{I}_{Gg} = [1, 1, 0, 1, 0, 1, 0, 0, 0, \ldots, 0]$. $\tilde{\mathbf{W}}_{Gg}$ represents the $g$th group.

We use Formula (14) to derive $\tilde{\mathbf{W}}$, and we can get the following formula.

$$-2\mathbf{X}\mathbf{Y}^T + 2\mathbf{X}\mathbf{X}^T\tilde{\mathbf{W}} + 2\alpha\mathbf{Q}\tilde{\mathbf{W}} + 2\beta\mathbf{F}\tilde{\mathbf{W}}. \tag{17}$$

Let the above formula be zero, and the solution of $\tilde{\mathbf{W}}$ can be obtained as follows.

$$\tilde{\mathbf{W}} = (\mathbf{X}\mathbf{X}^T + \alpha\mathbf{Q} + \beta\mathbf{F})^{-1}\mathbf{X}\mathbf{Y}^T. \tag{18}$$

The pseudo code of our algorithm is shown in Algorithm 1. The convergence conditions of the algorithm are as follows

$$\frac{|obj(t+1) - obj(t)|}{obj(t)} \leq 10^{-5}, \tag{19}$$

where $obj(t+1)$ is the value of the objective function in the $(t+1)$th iteration and $obj(t)$ is the value of the objective function in the $t$th iteration.

**Algorithm 1.** Pseudo Code for Proposed Method

**Input**: Training set $\mathbf{X} \in \mathbb{R}^{n\times d}$, Labels of the training data set $\mathbf{X}_{label} \in \mathbb{R}^{1\times n}$, Test Data $\mathbf{Y} \in \mathbb{R}^{m\times d}$;
**Output**: Label of test data;
Initialize $t$=1;
Randomly initialize $\mathbf{W}^{(0)}$;
Group the data by cosine similarity function or fuzzy C mean clustering to construct $\mathbf{I}_{Gg}$;
**repeat**
  Compute $\mathbf{F}^{(t+1)}$ via $f_{ii} = \sum_{g\in\Omega} \frac{(\mathbf{I}_{Gg})_i \|\tilde{\mathbf{W}}_{Gg}\|_1}{\|\tilde{\mathbf{W}}^i\|_1} (i = 1,\ldots,n)$;
  Compute $\mathbf{Q}^{(t+1)}$ via $q_{jj} = \frac{\sum_{v=1}^{n}\sqrt{\|\tilde{\mathbf{w}}^v\|_2^2}}{\sqrt{\|\tilde{\mathbf{w}}^j\|_2^2}}$;
  Updata $\mathbf{W}^{(t+1)}$ via Eq. (18);
  $t$ = $t$+1;
**until** *converge*
Get the label of the test data according to Formula (6);

### 3.4 Convergence Analysis

**Lemma 1.** *For any two non-negative vectors* $\mathbf{a} \in \mathbb{R}^{d\times 1}$ *and* $\mathbf{b} \in \mathbb{R}^{d\times 1}$*, the following formula holds.*

$$\left(\sum_{j=1}^{d} a_j\right)^2 - \sum_{j=1}^{d} b_j \sum_{j=1}^{d} \frac{{a_j}^2}{b_j} \leq \left(\sum_{j=1}^{d} b_j\right)^2 - \sum_{j=1}^{d} b_j \sum_{j=1}^{d} \frac{{b_j}^2}{b_j}. \tag{20}$$

**Proof.** According to the Cauchy-Schwarz inequality, the following formula can be obtained.

$$\left(\sum_{j=1}^{d} a_j\right)^2 = \left(\sum_{j=1}^{d} \frac{a_j}{\sqrt{b_j}}\sqrt{b_j}\right)^2 \leq \sum_{j=1}^{d} \frac{{a_j}^2}{b_j} \sum_{j=1}^{d} b_j. \tag{21}$$

Therefore, we can get

$$\left(\sum_{j=1}^{d} a_j\right)^2 - \sum_{j=1}^{d} \frac{{a_j}^2}{b_j} \sum_{j=1}^{d} b_j \leq \left(\sum_{j=1}^{d} b_j\right)^2 - \sum_{j=1}^{d} b_j \sum_{j=1}^{d} \frac{{b_j}^2}{b_j}. \tag{22}$$

□

**Theorem 2.** *In Algorithm 1, the value of the objective function (14) is monotonically decreasing in each iteration.*

**Proof.** By setting the value of $\mathbf{W}$ in the $t$th iteration to $\mathbf{W}^{(t)}$, Formula (14) can be written as follows.

$$\tilde{\mathbf{W}}^{(t+1)} = \arg\min \left\|\mathbf{Y}^T - \mathbf{X}^T\tilde{\mathbf{W}}\right\|_F^2 + \alpha tr(\tilde{\mathbf{W}}^T \mathbf{Q}^{(t)} \tilde{\mathbf{W}}) + \beta tr(\tilde{\mathbf{W}}^T \mathbf{F}^{(t)} \tilde{\mathbf{W}}). \tag{23}$$

We can get the following formula.

$$\begin{aligned}
&\left\|\mathbf{Y}^T - \mathbf{X}^T\tilde{\mathbf{W}}^{(t+1)}\right\|_F^2 \\
&+ \alpha \sum_{j=1}^{n} \sqrt{\left\|\tilde{\mathbf{W}}_j^{(t)}\right\|_2^2 + \varepsilon} \sum_{j=1}^{n} \frac{\left\|\tilde{\mathbf{W}}_j^{(t+1)}\right\|_2^2 + \varepsilon}{\sqrt{\left\|\tilde{\mathbf{W}}_j^{(t)}\right\|_2^2 + \varepsilon}} \\
&+ \beta tr\left((\tilde{\mathbf{W}}_j^{(t+1)})^T \sum_{j,g} \frac{(I_{Gg})_j \left\|\tilde{\mathbf{W}}_{Gg}^{(t)}\right\|_1}{\left\|\tilde{\mathbf{W}}_j^{(t)}\right\|_1} \tilde{\mathbf{W}}_j^{(t+1)}\right) \\
\leq &\left\|\mathbf{Y}^T - \mathbf{X}^T\tilde{\mathbf{W}}^{(t)}\right\|_F^2 \\
&+ \alpha \sum_{j=1}^{n} \sqrt{\left\|\tilde{\mathbf{W}}_j^{(t)}\right\|_2^2 + \varepsilon} \sum_{j=1}^{n} \frac{\left\|\tilde{\mathbf{W}}_j^{(t)}\right\|_2^2 + \varepsilon}{\sqrt{\left\|\tilde{\mathbf{W}}_j^{(t)}\right\|_2^2 + \varepsilon}} \\
&+ \beta tr\left((\tilde{\mathbf{W}}_j^{(t)})^T \sum_{j,g} \frac{(I_{Gg})_j \left\|\tilde{\mathbf{W}}_{Gg}^{(t)}\right\|_1}{\left\|\tilde{\mathbf{W}}_j^{(t)}\right\|_1} \tilde{\mathbf{W}}_j^{(t)}\right).
\end{aligned} \tag{24}$$

According to Lemma 1, we can get

$$\begin{aligned}
&\alpha \left(\sum_{j=1}^{n} \sqrt{\left\|\tilde{\mathbf{W}}_j^{(t+1)}\right\|_2^2 + \varepsilon}\right)^2 \\
&- \alpha \sum_{j=1}^{n} \sqrt{\left\|\tilde{\mathbf{W}}_j^{(t)}\right\|_2^2 + \varepsilon} \sum_{j=1}^{n} \frac{\left\|\tilde{\mathbf{W}}_j^{(t+1)}\right\|_2^2 + \varepsilon}{\sqrt{\left\|\tilde{\mathbf{W}}_j^{(t)}\right\|_2^2 + \varepsilon}} \\
\leq &\alpha \left(\sum_{j=1}^{n} \sqrt{\left\|\tilde{\mathbf{W}}_j^{(t)}\right\|_2^2 + \varepsilon}\right)^2 \\
&- \alpha \sum_{j=1}^{n} \sqrt{\left\|\tilde{\mathbf{W}}_j^{(t)}\right\|_2^2 + \varepsilon} \sum_{j=1}^{n} \frac{\left\|\tilde{\mathbf{W}}_j^{(t)}\right\|_2^2 + \varepsilon}{\sqrt{\left\|\tilde{\mathbf{W}}_j^{(t)}\right\|_2^2 + \varepsilon}}.
\end{aligned} \tag{25}$$

From the above two formulas, we can get

$$\begin{aligned}
&\left\|\mathbf{Y}^T - \mathbf{X}^T\tilde{\mathbf{W}}^{(t+1)}\right\|_F^2 + \alpha(\sum_{j=1}^{n} \sqrt{\left\|\tilde{\mathbf{W}}_j^{(t+1)}\right\|_2^2 + \varepsilon})^2 \\
&+ \beta tr((\tilde{\mathbf{W}}^{(t+1)})^T \mathbf{F}^{(t)} \tilde{\mathbf{W}}^{(t+1)}) \\
\leq &\left\|\mathbf{Y}^T - \mathbf{X}^T\tilde{\mathbf{W}}^{(t)}\right\|_F^2 + \alpha(\sum_{j=1}^{n} \sqrt{\left\|\tilde{\mathbf{W}}_j^{(t)}\right\|_2^2 + \varepsilon})^2 \\
&+ \beta tr((\tilde{\mathbf{W}}^{(\mathbf{t})})^T \mathbf{F}^{(t)} \tilde{\mathbf{W}}^{(t)}).
\end{aligned} \tag{26}$$

□

## 4 Experiments

In this part, we compare the classification performance of all algorithms on simulated data sets and UCI datasets.[1]

1. http://archive.ics.uci.edu/ml.

TABLE 2
Information About the Downloaded Datasets

| Datasets | Number of samples | Dimensions | Classes |
|---|---|---|---|
| Arrhythmia | 452 | 279 | 13 |
| Connection | 208 | 60 | 2 |
| Derma | 366 | 33 | 6 |
| Drift | 1244 | 129 | 5 |
| Ecoli | 336 | 343 | 8 |
| German | 1000 | 20 | 2 |
| Glass | 214 | 9 | 6 |
| Parkinson | 195 | 22 | 2 |
| Qsar | 1055 | 41 | 2 |
| Secom | 1567 | 590 | 2 |
| Spectf | 267 | 44 | 2 |
| Stalog | 6435 | 36 | 6 |

### 4.1 Datasets and Settings

In this section, we use 4 simulated datasets to verify the effectiveness of the proposed algorithm. Every simulated dataset contains redundant features and noise. Specifically, we first generate $n_i$ samples for the $i$th class through the multivariate normal distribution to obtain the data matrix $\mathbf{X}$. In the data matrix $\mathbf{X}$, its first $d$ columns are related to class labels, and the following $(d - d')$ columns are not related to class labels. i.e., the last $(d - d')$ columns are redundant features. Then we adjust the second parameter $\eta$ of the function $awgn$ in Matlab to increase the noise of different degrees. i.e., the larger the value of $\eta$, the more noise in $\mathbf{X}$. Finally, we generate class label $y(1, 2, \ldots, c)$, and standardize and centralize $\mathbf{X}$. Based on the above steps, 4 simulated datasets are generated, their detailed information is: $Data1(n_1 = 200, n_2 = 800,\ d = 60, d' = 40, \eta = 6.76)$, $Data2(n_1 = 500, n_2 = 500, d = 60, d' = 40, \eta = 33.8)$, $Data3(n_1 = 100, n_2 = 100, n_3 = 400, n_4 = 400, d = 60, d' = 40, \eta = 6.76)$, $Data4(n_1 = 250, n_2 = 250, n_3 = 250, n_4 = 250, d = 60, d' = 40, \eta = 33.8)$. In addition, we also used the UCI dataset to verify the performance of the algorithm. Including Spectf, Glass, German, Secom, Qsar, Parkinson, Ecoli, Derma, Arrhythmia, Stalog, Connectionist and Drift. Their information is shown in Table 2.

Arrhythmia is a data set of different types about arrhythmias. It includes 279 features, of which 206 are linear values and the rest are nominal values. The connection dataset contains 208 patterns, 111 of which are obtained from the rebound sonar signal on the metal cylinder. Under different conditions, the remaining 97 patterns are obtained from the rock. Derma is a data set for the differential diagnosis of erythematous and squamous skin diseases. It includes psoriasis, sebaceous dermatitis, lichen planus, pityriasis rosea, keratodermatitis and pityriasis rubra. Among these diseases, the initial characteristics of one disease are the characteristics of another, so it is not easy to be differentiated. Drift contains measurements of 16 chemical sensors exposed to six gases at different concentration levels. It was collected at the gas transfer platform facility from January 2008 to February 2011. Ecoli is a data set about the protein of Ecoli, which contains the information of amino acid content and cleavable sequence of protein. German is a data set about user credit. It includes information about the customer's checking account status, credit history, gender and property. Glass is a data about the classification of glass types. The classification of glass types is promoted by criminological investigation, and the glass left at the scene of the crime can be used as evidence to a certain extent. Parkinson consists of a series of biomedical speech measurements from 31 people, 23 of whom have Parkinson's disease. The QSAR data set includes the biodegradation experimental values of 1,055 chemicals. It can be used to study the relationship between chemical structure and biodegradation. It has 41 features and is a binary data set. Secom is a semiconductor manufacturing process data. It includes 1,567 samples that are described by 591 features. Spectf is a single proton emission computed tomography image of the heart. It includes information about two types of patients. Stalog is a data set about pixel classification in satellite images.

### 4.2 Comparied Algorithms

KNN [54]: It is the traditional KNN algorithm through euclidean distance and majority classification rules. Specifically, it first finds the K nearest neighbors of the test data from the training dataset according to the euclidean distance measurement. Then it takes the class with the highest frequency among the K neighbors as the class label of the test data. In our experiment, we set K values of 4, 5, and 6 and choose the best result as the final result.

W-KNN [55]: This method is an improved KNN algorithm, which applies a weighted classification rule. After obtaining the K neighbors of the test data, it sets the weight of each neighbor according to the distance. The closer the neighbor has the greater the weight, it has the greatest impact on the label of the test data. i.e., this method is different from traditional KNN (the voting weight of each neighbor is the same), it gives different voting weights to different neighbors.

CF-KNN [56]: This method is a two-step linear representation KNN algorithm. Specifically, it first uses all training samples to linearly represent a test sample to obtain the representation coefficient. Then it uses the representation coefficient vector and training data to fit the test data to get the error. At the same time, it selects the first $n_0$ smallest errors and training data to linearly represent the test data again. In this way, a new distance function is retained. Finally, K nearest neighbors are found through the new distance function and test data is classified according to the majority classification rules.

LMRKNN [13]: It first finds K neighbors from each class of training data, and constructs a local mean vector from the K neighbors of each class. Then the relationship between the local mean variable and the test data is constructed through linear representation, and the relationship matrix is obtained. Finally, the relationship matrix is used to construct a new distance measurement function, and the distance function is used to find K nearest neighbors to obtain the class label of the test data according to the majority rule.

DCLA-KNN [14]: This method is a locally adaptive K-nearest neighbor algorithm. The traditional KNN algorithm only considers the majority class, and it not only considers the majority class, but also considers the second majority class. i.e., it first selects the first majority class and the second majority class as the discrimination classes. Then it chooses an optimal K value according to the proposed rules and ordering. Finally, KNN classification is performed through the selected optimal K value. Its disadvantage is that the amount of calculation is relatively large, because when looking for the optimal K value, it is necessary to consider the situation under each K value.

K*tree [15]: This method is a non-lazy KNN algorithm. In the training process, it learns the optimal K value of each training sample and establishes the k*tree. In the test phase, it can find the optimal K value of each test data through the K*tree. Although it narrows the space range of searching neighbor, its K value calculation and neighbor searching are still carried out in two steps.

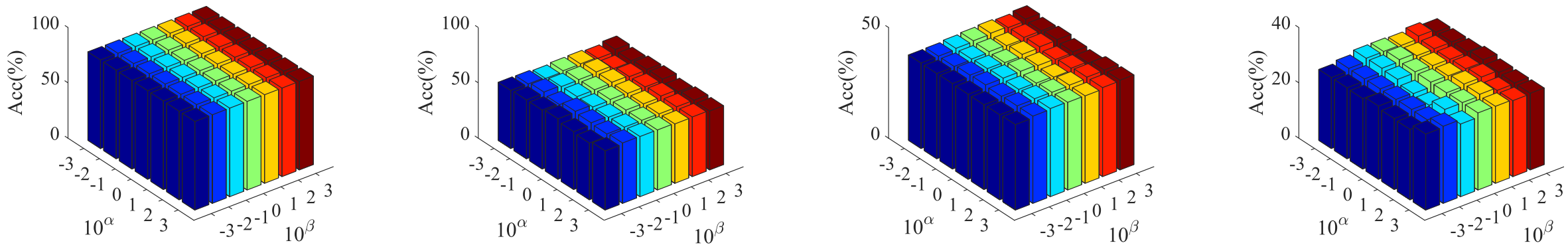

(a) Data1 (b) Data2 (c) Data3 (d) Data4

Fig. 2. The parameter sensitivity of the proposed algorithm on the simulated datasets.

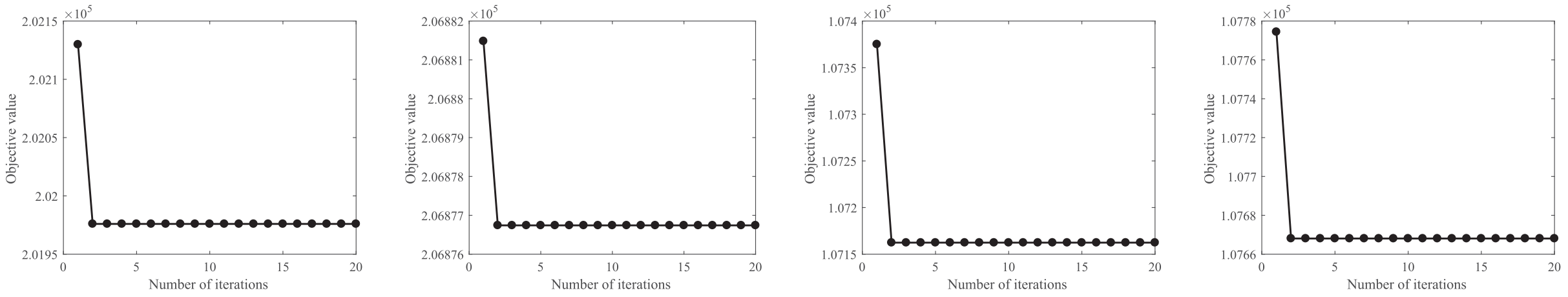

(a) Data1 (b) Data2 (c) Data3 (d) Data4

Fig. 3. The variations of the Objective Function Values (OFV) of our proposed method on the simulated datasets.

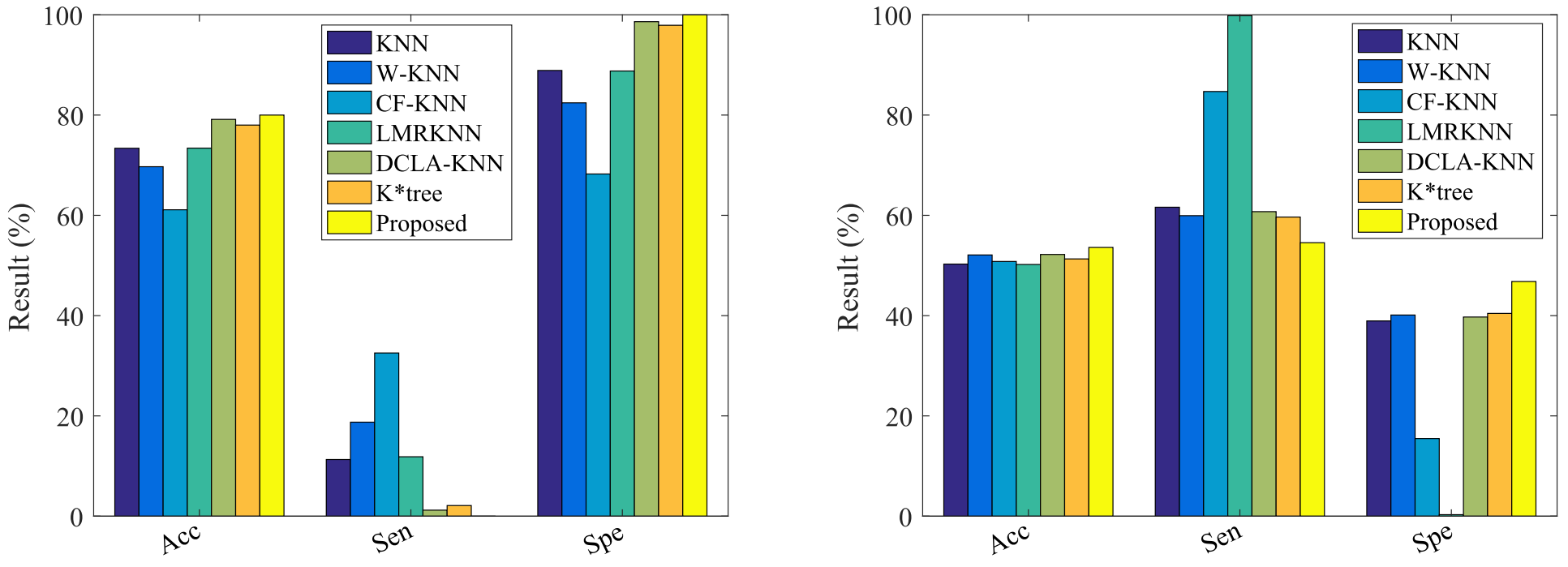

(a) Data1 (b) Data2

Fig. 4. Experimental results of all algorithms on the simulated binary dataset.

It should be noted that the code of the proposed algorithm is as follows: https://github.com/Lijy207/one-step-knn

### 4.3 Experimental Setting

This experiment is run on MATLAB R2016b, and the experimental platform is 64 bit win10 operating system. For the dataset, we made four simulated datasets, Data1 and Data2 are binary classification datasets, and Data3 and Data4 are multi-classification datasets. In the real dataset, there are 6 binary classification datasets and 6 multi-classification datasets.

KNN is used as the benchmark algorithm. W-KNN improves the KNN algorithm by proposing new classification rules. Both CFKNN and LMRKNN improve the KNN algorithm by proposing new distance functions. DCLA-KNN improves KNN by selecting the optimal K value.

In our experiment, we use 10-fold cross-validation to divide the dataset (i.e., It divides the data into 10 parts, 9 of which are used as the training set, and the remaining one as the test set) until all the data is used as the test set. In our proposed algorithm, we set the range of parameters as: $\alpha \in \{10^{-3}, 10^{-2}, 10^{-1}, 1, 10, 10^{2}, 10^{3}\}$ and $\beta \in \{10^{-3}, 10^{-2}, 10^{-1}, 1, 10, 10^{2}, 10^{3}\}$. In addition to the traditional KNN, for other comparison algorithms, we set appropriate parameters according to the corresponding references.

### 4.4 Experiments on Simulated Datasets

It should be noted that we set $\eta = 6.76$ in Data1 and Data3. It means that 1% noise is added to them. In the same way, 5% noise is added to Data2 and Data4. The results of all algorithms on the simulated data sets Data1, Data2, Data3, and Data4 are shown in Fig. 2, Fig. 4 and Table 3. From Table 3, we can see that in the Data1 and Data3 datasets, our proposed algorithm has achieved the best performance. Among them, compared with the oldest comparison algorithm KNN and the latest comparison algorithm DCLA-KNN, our proposed algorithm has improved by 5.46% and 0.22% respectively on the Data3 dataset. On the Data2 and Data4 datasets, the effects of all algorithms are not much different. It is due to the distribution of the simulated dataset itself, such as redundant features that affect the performance of the algorithm.

### 4.5 Experiments on Real Datasets

#### *4.5.1 Result Analysis*

No matter on the binary classification dataset or the multi-class dataset, our algorithms have achieved good results in most

TABLE 3
Accuracy ( mean ± standard deviation) for Simulated Datasets (%)

| Datasets | KNN | W-KNN | CF-KNN | LMRKNN | DCLA-KNN | K*tree | Proposed |
|---|---|---|---|---|---|---|---|
| Data1 | 73.38±0.01 | 69.70±0.01 | 61.11±0.01 | 73.40±0.02 | 79.14 ±0.01 | 78.00±0.01 | **80.00±0.01** |
| Data2 | 50.28±0.01 | 52.10±0.01 | 50.80±0.01 | 50.20±0.01 | 52.20±0.02 | 51.31±0.02 | **53.60±0.02** |
| Data3 | 34.54±0.01 | 34.98±0.01 | 30.88±0.02 | 38.74±0.02 | 39.78±0.02 | 40.4±0.01 | **40.00±0.02** |
| Data4 | 26.70±0.02 | 25.00±0 | 26.30±0.02 | 25.20±0.01 | 25.20±0.01 | 27.30±0.12 | **27.70±0.02** |

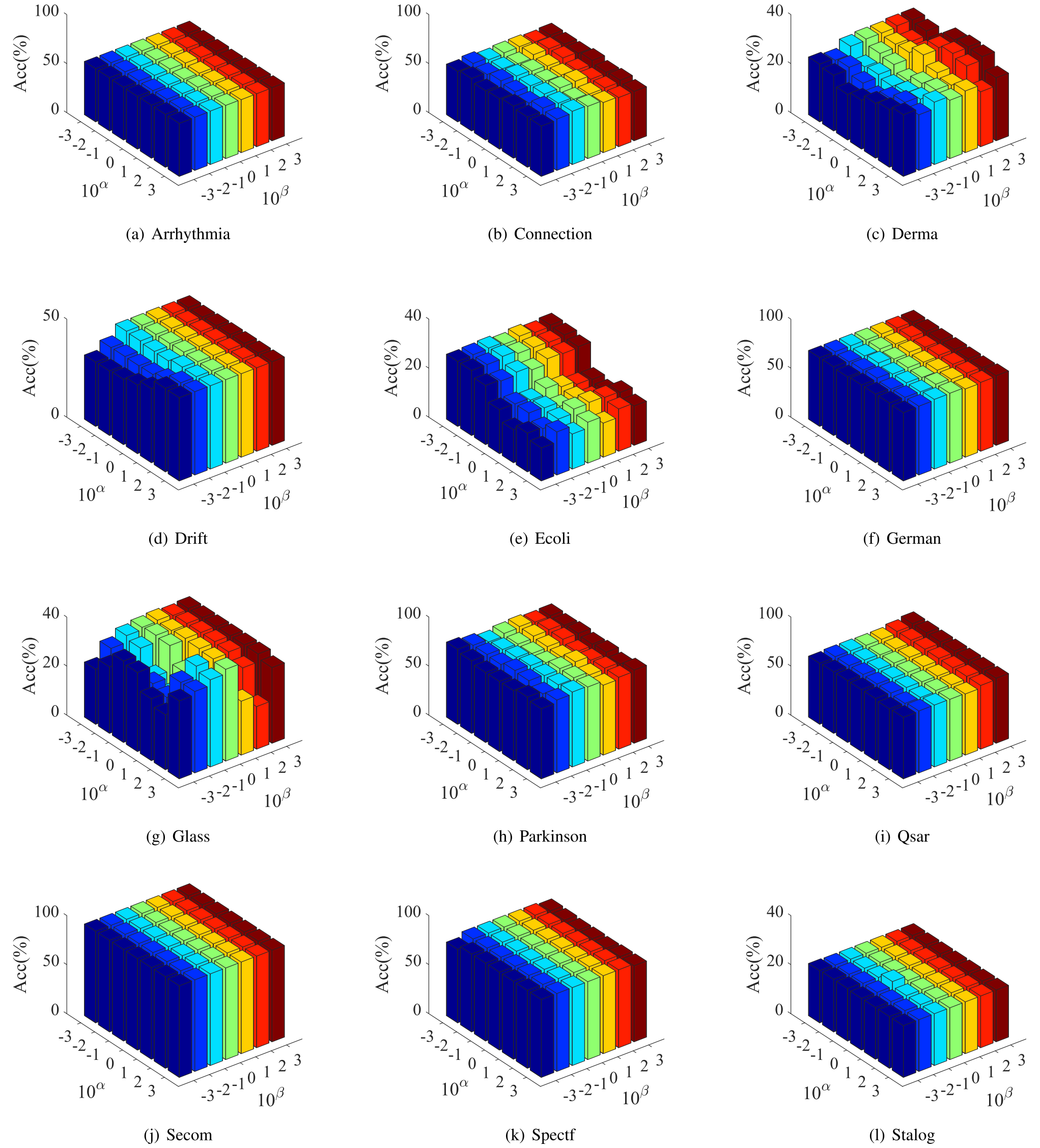

(a) Arrhythmia (b) Connection (c) Derma
(d) Drift (e) Ecoli (f) German
(g) Glass (h) Parkinson (i) Qsar
(j) Secom (k) Spectf (l) Stalog

Fig. 5. The parameter sensitivity of the proposed algorithm on the UCI datasets.

cases. As shown in Table 5 and Fig. 7. Specifically, compared with the oldest comparison algorithm KNN and the latest comparison algorithm DCLA-KNN on all datasets, the proposed algorithm has an average increase of 7.24% and 4.16%, respectively. Because the proposed algorithm improves KNN from a new perspective. The proposed objective function (i.e., Formula (4)) can not only get the optimal K value of each test sample, but also the corresponding K neighbors. To a certain extent, it can be regarded as a new measurement function. In addition, the proposed algorithm also applies a new weighted classification rule, which is not a simple distance-based weighting, it is constructed by the relationship between test samples and training samples. In fact, W-KNN focuses on classification rules, CFKNN and LMRKNN focus on new distance measurement functions, and DCLA-KNN focuses on the selection of K values. That is, they are only considered from the 1 challenge of KNN, and our proposed algorithm not only considers the optimal K value, the nearest neighbor query, but also considers the classification rules.

### 4.5.2 Parameter Sensitivity

As shown in Equation (4), our proposed algorithm has two parameters, namely $\alpha$ and $\beta$. Both $\alpha$ and $\beta$ control the weight of neighbors. And $\beta$ also controls the sparsity of $\mathbf{W}$ to a certain extent, and it affects the number of neighbors of the test data.

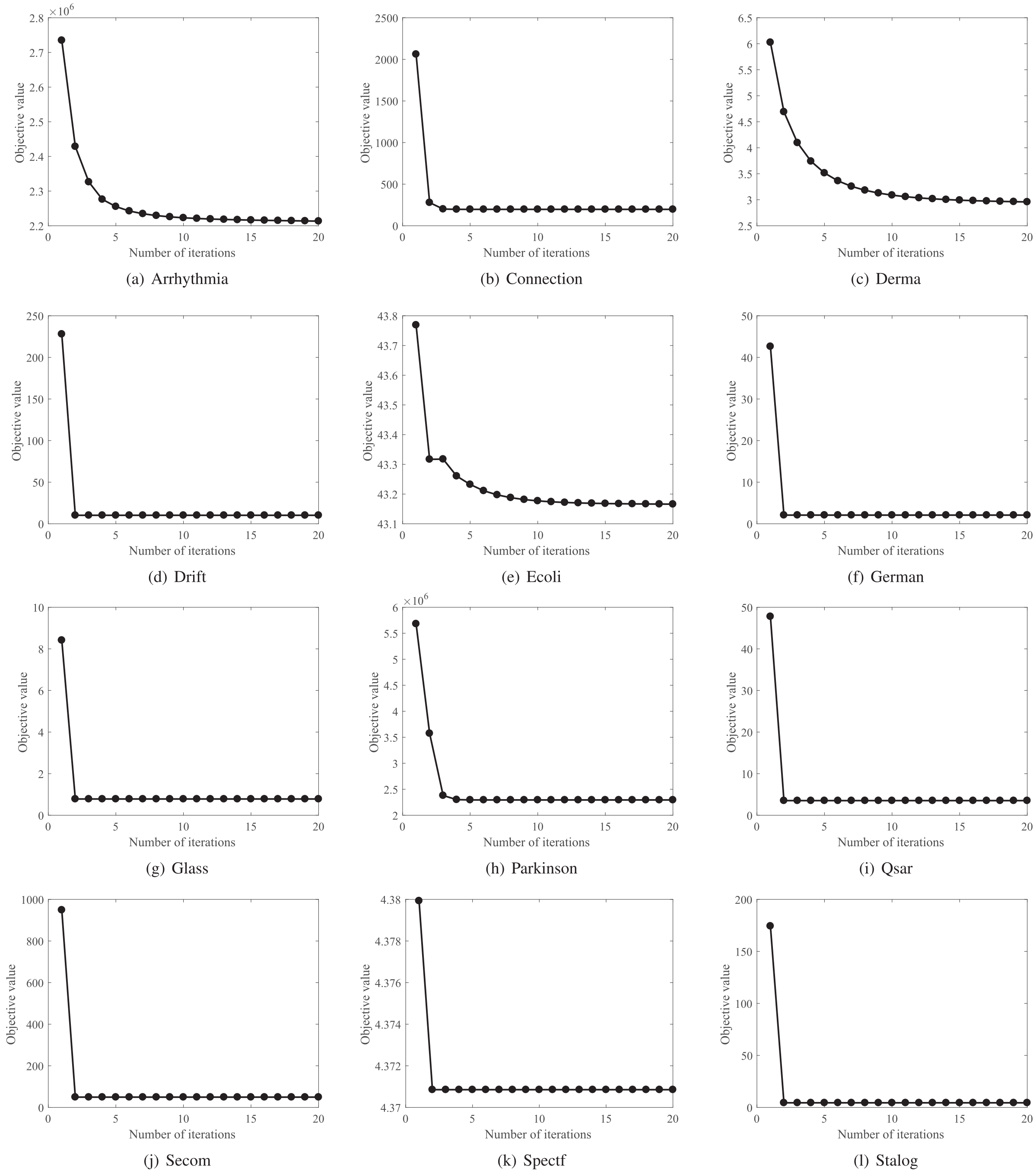

Fig. 6. The variations of the Objective Function Values (OFV) of our proposed method on the UCI datasets.

It can be seen from the Fig. 5 that our proposed algorithm is more sensitive to parameters $\alpha$ and $\beta$. Therefore, we need to carefully adjust the values of their two parameters.

#### 4.5.3 Convergence Analysis

In the experiment, the change in the value of the proposed objective function in each iteration is shown in Figs. 3 and 6. We set the termination condition of the iteration to be $\frac{|obj(t+1)-obj(t)|}{obj(t)} \leq 10^{-5}$. As can be seen from Figs. 3 and 6, our objective function value gradually decreases as the number of iterations increases, until convergence. In most data sets, the objective function converges within 15 iterations, which shows that the proposed algorithm converges quickly.

#### 4.5.4 Running Cost Analysis

Tables 4 and 6 show the running cost of all algorithms on simulated dataset and real dataset respectively. From Tables 4 and 6, we can see that the running cost of KNN, W-KNN, K*tree and the proposed algorithm are relatively low. The running cost of CF-KNN, LMRKNN and DCLA-KNN is higher.

Because traditional KNN is a lazy learning algorithm, it does not do any model training, the training time cost is 0. It only needs to classify according to the known K value, distance function and major classification rules. W-KNN is an improvement on the classification rules. It will take more time to calculate weight than traditional KNN. Both CF-knn and LMRknn improve KNN by constructing new distance measures. They need to spend more time to construct new distance functions, except for classification through the majority rule. DCLA-KNN needs to compute not only the major class, but also the second major class. In addition, it has to choose the optimal K value. So it takes more time. K*tree includes the training part and the test part. In the training process, it needs to obtain the optimal K value and subset of training data, i.e., build K*tree. In the test process, it uses the majority rule to classify the test data according to K*tree. It should be noted that in reference [15] (K*tree), the experiment compares the running cost of the test phase of K*tree. Because K*tree reduces the search range, compared with the traditional KNN, the running cost is low. However, in this paper, we compare the running cost of setting K and completely searching k nearest neighbors, so the running cost of K*tree is higher than KNN.

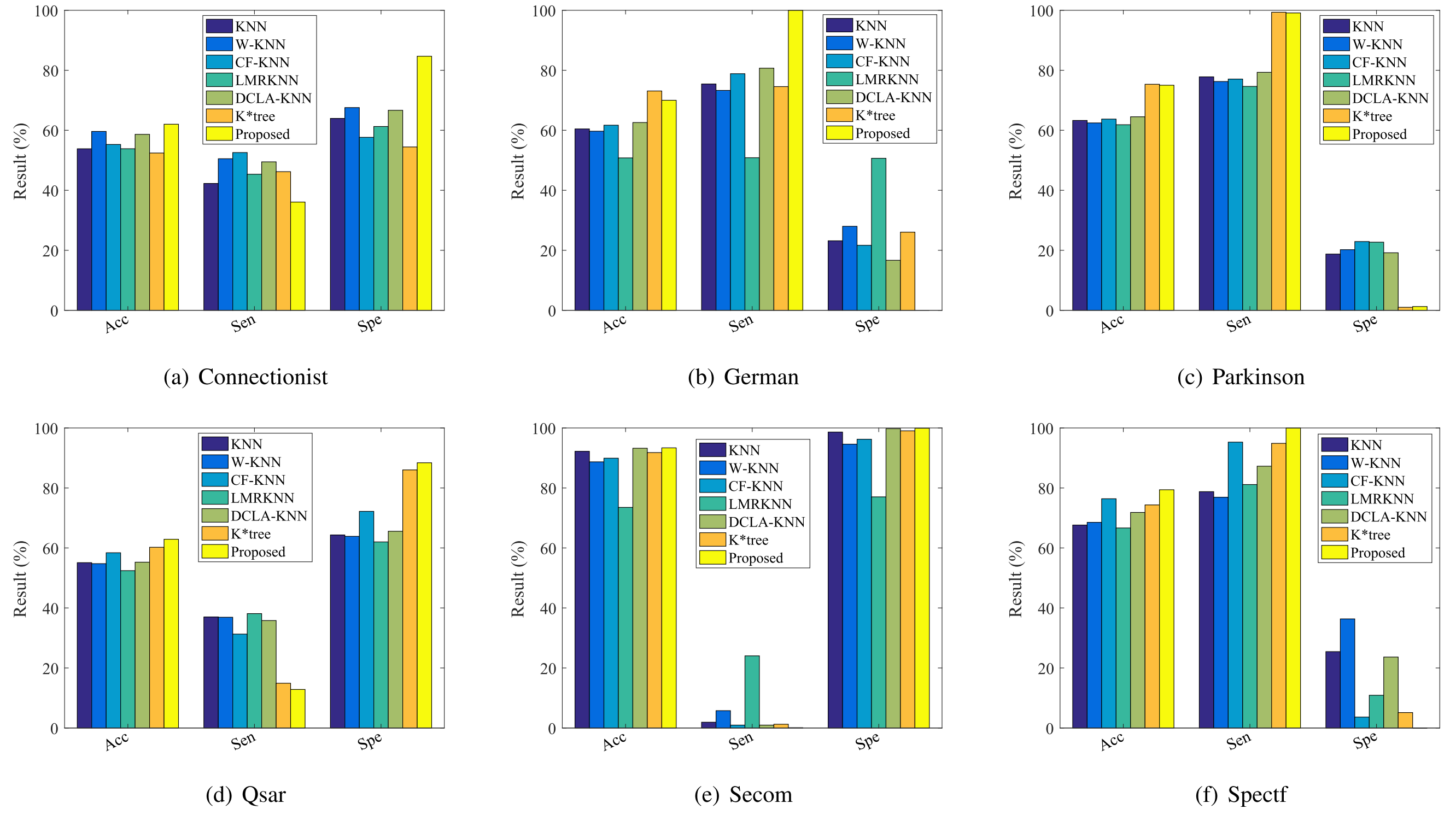

(a) Connectionist (b) German (c) Parkinson

(d) Qsar (e) Secom (f) Spectf

Fig. 7. Experimental results of all algorithms on the UCI binary dataset.

TABLE 4
The Running Cost of all Algorithms on the Simulated Datasets (second)

| Datasets | KNN | W-KNN | CF-KNN | LMRKNN | DCLA-KNN | K*tree | Proposed |
|---|---|---|---|---|---|---|---|
| Data1 | 0.07 | 0.67 | 146.98 | 1.19 | 43.94 | 1.29 | **0.04** |
| Data2 | 0.05 | 0.70 | 103.68 | 1.56 | 27.95 | 1.15 | **0.03** |
| Data3 | 0.04 | 0.50 | 88.65 | 1.91 | 26.06 | 1.14 | **0.06** |
| Data4 | 0.04 | 0.53 | 105.85 | 1.46 | 21.18 | 0.97 | **0.05** |

For the proposed method, in the training process, we need to get a corresponding relationship vector for each test data. In the test process, KNN classification only needs to be carried out according to the obtained optimal K value and the corresponding k-nearest neighbor. So, the training complexity of the proposed algorithm is $O(n^2)$, and in the test phase, the time complexity is $O(1)$. For K*tree method, its training time complexity is also $O(n^2)$. In the test phase, it only needs to find k nearest neighbors from the subset of training data according to K*tree. Therefore, the test time complexity is $O(\log(d) + s)$, where $s \ll n$.

TABLE 5
Accuracy ( mean ± standard deviation) for UCI Datasets (%)

| Datasets | KNN | W-KNN | CF-KNN | LMRKNN | DCLA-KNN | K*tree | Proposed |
|---|---|---|---|---|---|---|---|
| Arrhythmia | 34.95±0.01 | 37.17±0.01 | 33.21±0.02 | 36.59±0.01 | 47.57±0.01 | 78.50±0.02 | **54.20±0.01** |
| Connection | 51.59±0.02 | 53.07±0.02 | 52.07±0.03 | 52.40±0.02 | 52.93 ±0.02 | 52.43±0.01 | **54.04±0.02** |
| Derma | 19.51±0.03 | 16.72±0.01 | 22.02±0.02 | 19.67±0.02 | 20.49±0.02 | 21.70±0.01 | **22.90±0.02** |
| Drift | 27.41±0.01 | 26.85±0.01 | 34.49±0.01 | 26.37±0.02 | 30.28±0.01 | 32.68±0.03 | **34.68±0.01** |
| Ecoli | 29.94±0.01 | 30.95±0.01 | 26.19±0.01 | 34.79±0.02 | 31.90 ±0.02 | 45.60±0.01 | **32.26±0.03** |
| German | 60.75±0.01 | 59.17±0.02 | 61.23±0.02 | 56.67±0.02 | 62.82 ±0.01 | 73.10±0.03 | **70.00±0.01** |
| Glass | 28.60±0.02 | 35.23±0.01 | 25.23±0.03 | 32.24±0.01 | 31.73±0.02 | 35.41±0.01 | **36.50±0.09** |
| Parkinson | 63.29±0.02 | 65.13±0.02 | 63.74±0.04 | 65.64±0.03 | 66.15 ±0.02 | 75.34±0.02 | **75.03±0.01** |
| Qsar | 55.36±0.01 | 54.78±0.01 | 57.43±0.01 | 56.78±0.01 | 57.35±0.01 | 60.23±0.01 | **62.90±0.01** |
| Secom | 91.77±0.01 | 89.92±0.01 | 90.36±0.01 | 75.37±0.01 | 92.99±0.01 | 91.80±0.02 | **93.36±0.01** |
| Spectf | 68.28±0.03 | 67.42±0.03 | 77.68±0.01 | 73.41±0.01 | 72.17 ±0.03 | 74.35±0.02 | **79.40±0.01** |
| Stalog | 18.01±0.01 | 20.92±0.01 | 20.37±0.02 | 17.13±0.01 | 20.16±0.02 | 21.02±0.01 | **21.10±0.02** |

TABLE 6
The Running Cost of all Algorithms on the UCI Datasets (second)

| Datasets | KNN | W-KNN | CF-KNN | LMRKNN | DCLA-KNN | K*tree | Proposed |
|---|---|---|---|---|---|---|---|
| Arrhythmia | 0.13 | 0.31 | 12.53 | 2.91 | 22.23 | 0.14 | **0.09** |
| Connection | 0.04 | 0.10 | 1.42 | 0.19 | 6.18 | 0.24 | **0.01** |
| Derma | 0.03 | 0.17 | 4.82 | 0.98 | 8.31 | 0.29 | **0.02** |
| Drift | 0.05 | 1.21 | 165.39 | 3.54 | 45.65 | 1.21 | **0.07** |
| Ecoli | 0.03 | 0.20 | 5.14 | 1.30 | 22.92 | 0.47 | **0.02** |
| German | 0.63 | 0.79 | 80.53 | 1.08 | 20.89 | 0.13 | **0.05** |
| Glass | 0.09 | 0.14 | 0.76 | 0.47 | 3.47 | 0.24 | **0.02** |
| Parkinson | 0.02 | 0.08 | 1.05 | 0.17 | 4.21 | 0.15 | **0.01** |
| Qsar | 0.03 | 0.47 | 99.56 | 1.01 | 25.96 | 0.47 | **0.04** |
| Secom | 0.16 | 9.47 | 496.51 | 17.78 | 250.44 | 1.25 | **0.07** |
| Spectf | 0.02 | 0.13 | 4.53 | 0.28 | 9.54 | 0.27 | **0.01** |
| Stalog | 0.67 | 7.57 | 6411.3 | 23.18 | 324.96 | 3.75 | **2.04** |

Among them, W-KNN only increases the weight of the nearest neighbor on the basis of KNN, so their time cost is similar. The k*tree and proposed algorithm are both non lazy algorithms. The running cost of the proposed algorithm is the minimum, because the algorithm is one-step calculation in the training process. It can get the optimal K value and k-nearest neighbor of test data at the same time. In the test process, it only needs to perform neighbor label aggregation. However, the k*tree algorithm needs to find the nearest neighbor in the leaf node of the tree.

## 5 Conclusion

This paper has proposed a one-step KNN classification algorithm that integrates both setting K value and searching K nearest neighbors to a computation. Specially speaking, in the training process, we use the least square loss, $l_{2,1}^2$-norm and $l_1^2$-norm to obtain the relationship matrix and weight vector between the test data and the training data. In the testing process, we use the obtained relationship matrix and weight vector to classify the test data through weighted classification rules. We have conducted a series of experiments on simulated data sets and UCI data sets, which show that the proposed algorithm exceeds the state-of-the-art methods in terms of ACC and running cost.

As we all know, with the explosive growth of data, there is the problem of “dimension disaster”. The applicability of euclidean distance is greatly reduced by high-dimensional data. Therefore, in the future work, we plan to improve KNN from the following two aspects as follows.

1) We use a new metric function to adapt to high-dimensional data and improve the applicability of KNN.
2) The feature selection algorithm is embedded into KNN. The result of feature selection (dimensionality reduction) acts on KNN, and the classification result of KNN has a reaction on feature selection.

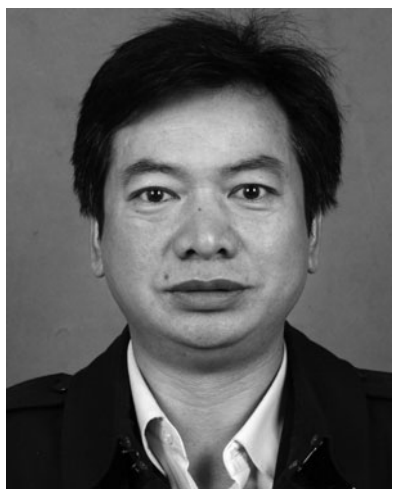

**Shichao Zhang** (Senior Member, IEEE) received the PhD degree from Deakin University, Australia. He is a China National Distinguished professor with Central South University, China. His research interests include data mining and big data. He has published 90 international journal papers and more than 70 international conference papers. He is a CI for 18 competitive national grants. He serves/served as an associate editor for four journals.

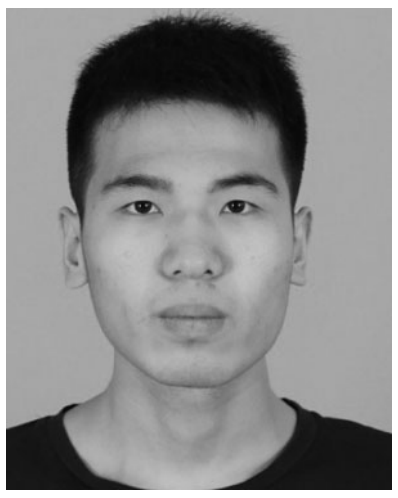

**Jiaye Li** is currently working toward the PhD degree at Central South University, China. His research interests include machine learning, data mining, and deep learning.